\documentclass[10pt,twocolumn,letterpaper]{article}
\usepackage{cvpr}             

\usepackage{graphicx}
\usepackage{amsmath}
\usepackage{amssymb}
\usepackage{booktabs}
\usepackage{bbding}
\usepackage{enumitem}
\usepackage[ruled]{algorithm2e}
\usepackage[dvipsnames]{xcolor}

\usepackage{multirow}
\usepackage{amsthm}
\usepackage{mathrsfs}
\usepackage{diagbox}
\usepackage{bm}
\usepackage{newfloat}
\usepackage{xspace}
\usepackage{xcolor}

\definecolor{citecolor}{RGB}{0, 113, 188}
\usepackage[pagebackref,breaklinks,colorlinks, citecolor=citecolor]{hyperref}

\usepackage[capitalize]{cleveref}
\crefname{section}{Sec.}{Secs.}
\Crefname{section}{Section}{Sections}
\Crefname{table}{Table}{Tables}
\crefname{table}{Tab.}{Tabs.}
\newcommand{\system}{ISVOS\xspace}

\begin{document}

\title{Look Before You Match: Instance Understanding Matters \\ in Video Object Segmentation}

\author{Junke Wang$^{1,2}$,~Dongdong Chen$^{3}$,~Zuxuan Wu$^{1,2}$,~Chong Luo$^{4}$,~Chuanxin Tang$^{4}$, \\ ~Xiyang Dai$^{3}$, ~Yucheng Zhao$^{4}$,~Yujia Xie$^{3}$,~Lu Yuan$^{3}$, ~Yu-Gang Jiang$^{1,2}$
\\
$^{1}$Shanghai Key Lab of Intell. Info. Processing, School of CS, Fudan University \\
$^{2}$Shanghai Collaborative Innovation Center on Intelligent Visual Computing \\
$^{3}$Microsoft Cloud + AI, $^{4}$Microsoft Research Asia 
}

\maketitle

\begin{abstract}
Exploring dense matching between the current frame and past frames for long-range context modeling, memory-based methods have demonstrated impressive results in video object segmentation (VOS) recently. Nevertheless, due to the lack of instance understanding ability, the above approaches are oftentimes brittle to large appearance variations or viewpoint changes resulted from the movement of objects and cameras. In this paper, we argue that instance understanding matters in VOS, and integrating it with memory-based matching can enjoy the synergy, which is intuitively sensible from the definition of VOS task, \ie, identifying and segmenting object instances within the video. Towards this goal, we present a two-branch network for VOS, where the query-based instance segmentation (IS) branch delves into the instance details of the current frame and the VOS branch performs spatial-temporal matching with the memory bank. We employ the well-learned object queries from IS branch to inject instance-specific information into the query key, with which the instance-augmented matching is further performed. In addition, we introduce a multi-path fusion block to effectively combine the memory readout with multi-scale features from the instance segmentation decoder, which incorporates high-resolution instance-aware features to produce final segmentation results. Our method achieves state-of-the-art performance on DAVIS 2016/2017 val (92.6\% and 87.1\%), DAVIS 2017 test-dev (82.8\%), and YouTube-VOS 2018/2019 val (86.3\% and
86.3\%), outperforming alternative methods by clear margins. 
\end{abstract}

\section{Introduction}
\label{sec:intro}
\begin{figure}[t]
  \centering
   \includegraphics[width=\linewidth]{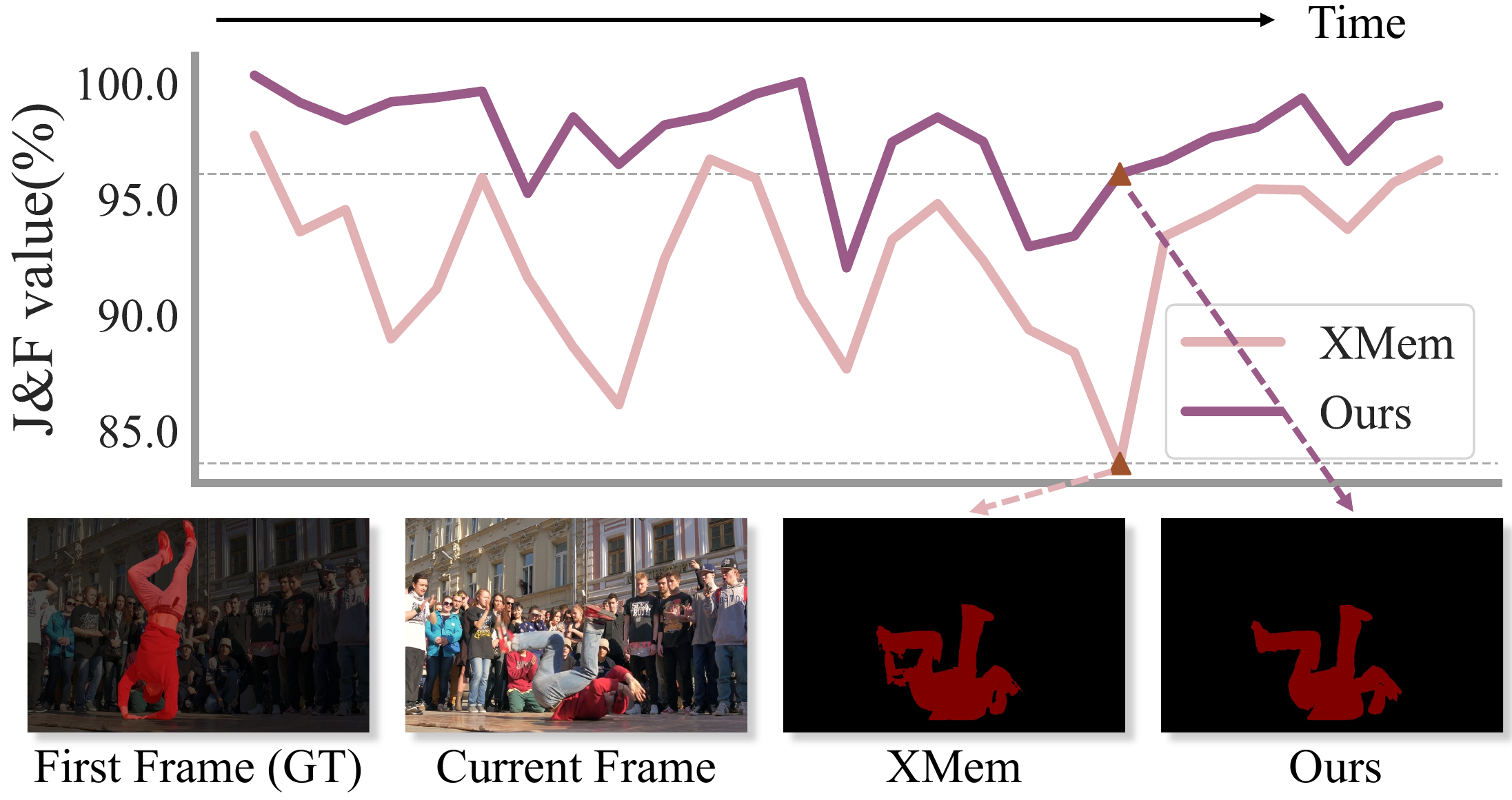}
   \vspace{-0.15in}
   \caption{J$\&$F-time curve of XMem~\cite{cheng2022xmem}, a state-of-the-art memory-based VOS model and our proposed method. XMem will suffer from a distinct accuracy degradation when the appearance of the target object (\eg, pose of the dancing person) changes dramatically compared to the reference frame. Comparatively, our approach is more robust to this challenging case.}
   \label{fig:badcase}
\end{figure}

Video object segmentation aims to identify and segment specific objects in a video sequence, which has very broad applications, \eg, interactive video editing and autonomous driving. This work focuses on the semi-supervised setting where the annotation of the first frame is given. Starting from Space-Time Memory network (STM)~\cite{oh2019video}, memory-based methods~\cite{liang2020video,lu2020video,cheng2021mivos,hu2021learning,seong2021hierarchical,cheng2021stcn,wang2021swiftnet,park2022per,cheng2022xmem} have almost dominated this field due to their superior performance and simplicity. STM~\cite{oh2019video} and its variants~\cite{li2020fast,xie2021efficient,hu2021learning} typically build a feature memory to store the past frames as well as corresponding masks, and perform dense matching between the query frame and the memory to separate targeted objects from the background.

Despite the prominent success achieved, there exists a non-negligible limitation for the above approaches, \ie, the object deformation and large appearance variations resulted from the motion of camera and objects will inevitably give rise to the risk of false matches~\cite{oh2019video,cheng2021stcn,cheng2022xmem}, thus making them struggle to generate accurate masks. We visualize the $J\&F$-time curve of XMem~\cite{cheng2022xmem}, a state-of-the-art memory-based VOS model, on a representative video from DAVIS 2017 in Figure~\ref{fig:badcase}. It can be seen that, when the target object undergoes a distinct pose change compared to the first reference frame, XMem~\cite{cheng2022xmem} misidentifies another person wearing the same color as the foreground and suffers from drastic performance degradation. 

\begin{figure}[t]
  \centering
   \includegraphics[width=\linewidth]{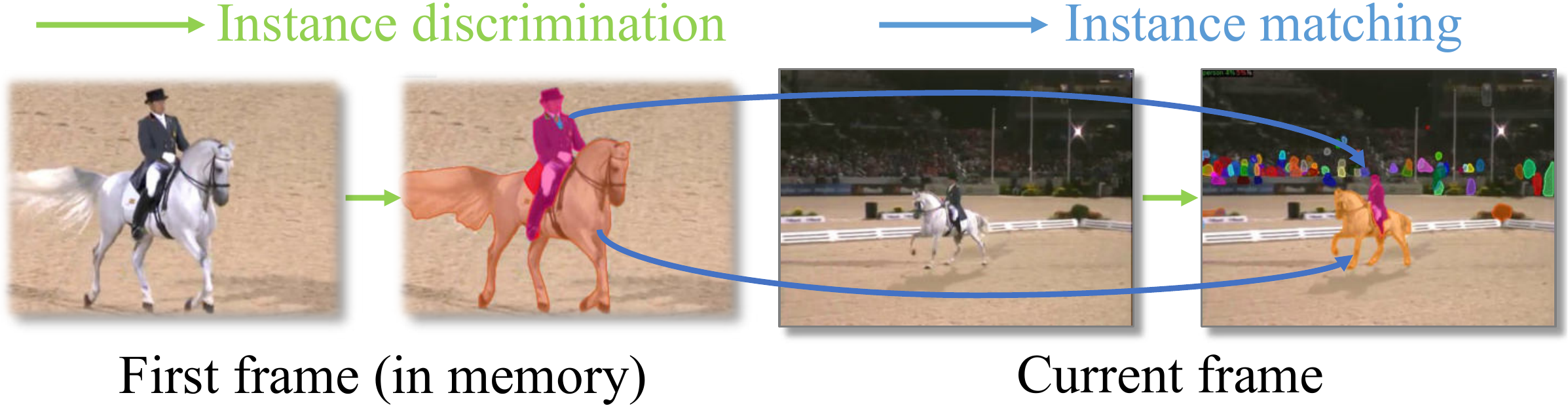}
   \vspace{-0.25in}
   \caption{\textbf{A conceptual introduction on how humans address the VOS task}. For the current frame of a video stream, humans first distinguish between different instances and then match them with the target object(s) in memory. }
   \label{fig:conceptual}
\end{figure}

In contrast, humans are capable of avoiding such mistakes and achieving consistently accurate matching. This gap motivates us to reflect on how we humans resolve the VOS task. Intuitively, given the current frame in a video sequence, humans typically first distinguish between different instances within it by identifying which instance each pixel belongs to. After that, the instance matching with the target object(s) in memory is conducted to obtain the final results (see Figure~\ref{fig:conceptual} for a conceptual illustration).  In fact, this intuition is also consistent with the definition of VOS itself, \ie, identify (matching) and segmenting objects (instance understanding). Moreover, in the absence of instance understanding, it is theoretically difficult to generate accurate predictions for regions that are invisible in reference frame by pure matching.

Inspired by this, we argue that instance understanding is critical to the video object segmentation, which could be incorporated with memory-based matching to enjoy the synergy. More specifically, we aim to derive instance-discriminative features that are able to \textbf{\textit{distinguish different instances}}. Equipped with these features, we then perform \textbf{\textit{semantic matching}} with the memory bank to effectively associate the target object(s) with specific instance(s) in the current frame.

In this spirit, we present a two-branch network, \system, for semi-supervised VOS, which contains an instance segmentation (IS) branch to delve into the instance details for the current frame and a video object segmentation branch that resorts to an external memory for spatial-temporal matching. The IS branch is built upon a query-based instance segmentation model~\cite{cheng2022masked} and supervised with instance masks to learn instance-specific representations. Note that \system is a generic framework and IS branch can be easily replaced with more advanced instance understanding models.
The video object segmentation (VOS) branch, on the other hand, maintains a memory bank to store the features of past frames and their predictions. We compare the query key of current frame and memory key\footnote{In this paper, we follow previous work~\cite{oh2019video,cheng2021stcn} of which are compared with query key to denote the key features of current frame and memory bank as query key and memory key, respectively, so as to perform instance-augmented matching.} from memory bank for affinity calculation following~\cite{oh2019video,seong2021hierarchical,cheng2021stcn,cheng2022xmem}. Motivated by recent approaches that use learnable queries serving as region proposal networks to identify instances in images~\cite{fang2021instances,cheng2021per,yang2021tracking,cheng2022masked}, we employ object queries from the IS branch to inject instance-specific information into our query key, with which the instance-augmented matching is performed. After that, the readout features are produced by aggregating the memory value with the affinity matrix. Moreover, in order to make use of the fine-grained instance details reserved in high-resolution instance-aware features, we further combine the multi-scale features from instance segmentation decoder with the memory readout through a carefully designed multi-path fusion block to finally generate the segmentation masks.

We conduct experiments on the standard DAVIS~\cite{perazzi2016benchmark,pont20172017} and YouTube-VOS~\cite{xu2018YouTube} benchmarks. The results demonstrate that our \system can achieve state-of-the-art performance on both single-object (\ie, 92.6\% in terms of $J\&F$ on DAVIS 2016 validation split) and multi-object benchmarks (\ie, 87.1\% and 82.8\% on DAVIS 2017 validation and test-dev split, 86.3\% and 86.3\% on YouTube-VOS 2018 $\&$ 2019 validation split) without post-processing. 

\section{Related Work}
\label{sec:related}
\noindent \textbf{Propagation-based VOS.} Propagation-based VOS methods~\cite{tsai2016video,cheng2017segflow,li2018video,oh2018fast,xu2018dynamic,yang2018efficient,duke2021sstvos,xu2022accelerating} take advantage of temporal correlations between adjacent frames to iteratively propagate the segmentation masks from the previous frame to the current frame. Early approaches~\cite{perazzi2017learning,caelles2017one,hu2017maskrnn} typically follow an online learning manner by finetuning models at test-time, which therefore suffer from limited inference efficiency. To mitigate this issue, the following studies shift attention to offline learning by utilizing optical flow~\cite{tsai2016video,cheng2017segflow,xu2018dynamic} as guidance to deliver the temporal information smoothly. Despite the promising results achieved, these methods are oftentimes vulnerable to the error accumulation brought by occlusion or drifting. 

\begin{figure*}[t]
  \centering
   \includegraphics[width=\linewidth]{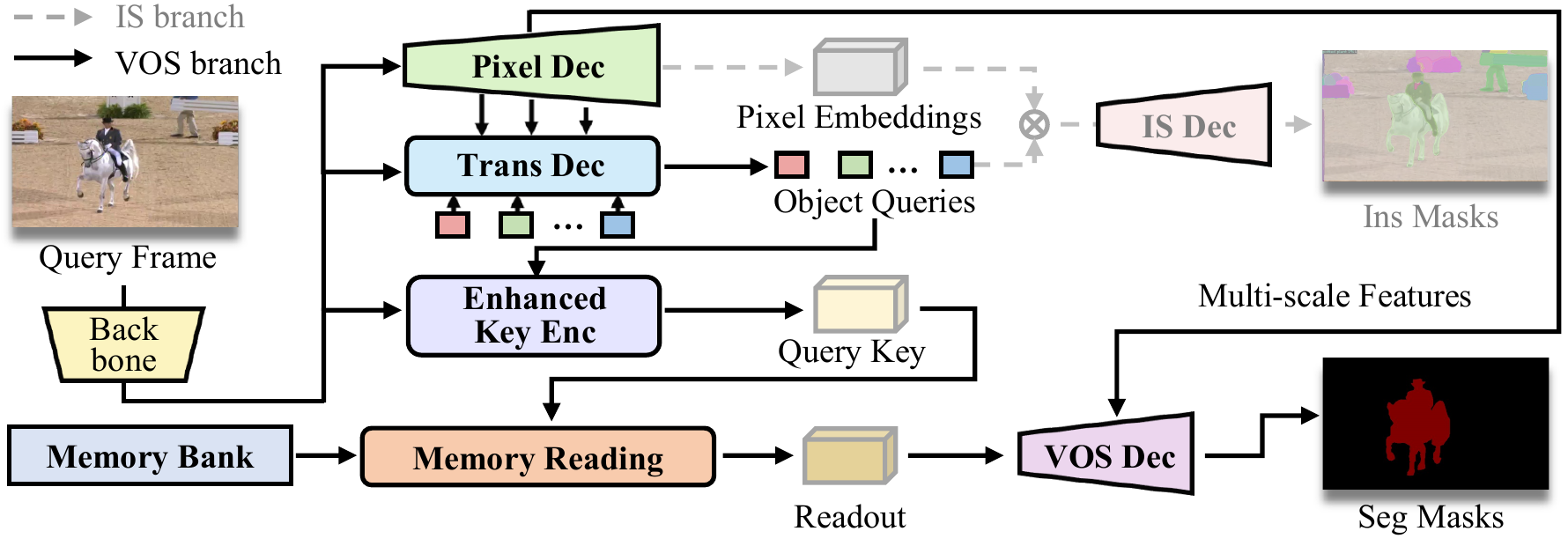}
   \vspace{-0.25in}
   \caption{Overview of the proposed method, which consists of an \textbf{\textcolor{gray}{instance segmentation branch}} and a \textbf{video object segmentation branch}. We jointly train both branches on instance segmentation and video object segmentation tasks, respectively. The IS branch parts denoted as dotted gray line will be skipped during inference, meaning that our method does not explicitly use the output instance masks.}
   \label{fig:network}
\end{figure*}

\vspace{0.02in}
\noindent \textbf{Matching-based VOS.} In order to model the spatial-temporal context over longer distances, matching-based models~\cite{hu2018videomatch,oh2019video,cheng2021mivos,cheng2021stcn} typically calculate the correspondence between the current frame and the reference frame~\cite{hu2018videomatch,chen2018blazingly,voigtlaender2019feelvos,yang2020collaborative} and even a feature memory~\cite{hu2021learning,liang2020video,lu2020video,seong2020kernelized,seong2021hierarchical,xie2021efficient,cheng2022xmem} to identify target objects. In addition, several studies focus on designing novel memory construction~\cite{li2020fast,liang2020video} or matching~\cite{wang2021swiftnet} strategies to improve the inference efficiency of VOS models. However, the widely adopted dense matching between reference features will inevitably fail on the objects with significant appearance variations or viewpoint changes. In this paper, we propose to integrate instance understanding into VOS, which is neglected in existing matching-based methods.

\vspace{0.02in}
\noindent \textbf{Instance Segmentation.} Built upon powerful object detectors, two-stage instance segmentation models~\cite{he2017mask,li2017fully,liu2018path,huang2019mask,bolya2019yolact,chen2020blendmask} predict bounding boxes first and then extract the instance mask in each region of interest (ROI). These methods require complicated procedures, \eg, region proposal generation~\cite{ren2015faster} and ROIAlign~\cite{he2017mask}, which motivates the following work to develop one-stage box-free models~\cite{arnab2017pixelwise,newell2017associative,liu2017sgn,kong2018recurrent,gao2019ssap,neven2019instance,cheng2020panoptic,wang2020solo,wang2020solov2}. Recently, the success of DETR~\cite{carion2020end} inspires a series of query-based models~\cite{fang2021instances,cheng2021per,cheng2022masked} to reformulate the instance segmentation from a novel ``set prediction'' perspective, which achieve state-of-the-art performance on standard benchmarks~\cite{lin2014microsoft}. In this work, we build our VOS model with an auxiliary query-based instance segmentation model~\cite{cheng2022masked} to make the intermediate feature instance-aware rather than explicitly use the output segmentation mask. It enables us to perform instance-augmented matching between the instance-aware features of the current frame and memory. Note that our approach is generic and can be employed to improve various memory-based VOS methods.

\section{Method}
\label{sec:method}
Our goal is to integrate the instance understanding for improved memory-based video object segmentation. To this end, we propose \system, a two-branch network where the instance segmentation (IS) branch learns instance-specific information, and the video object segmentation (VOS) branch performs instance-augmented matching through memory reading. Such a design also shares the similar spirit with a prior study~\cite{goodale1992separate} which implies that two layers of human cortex responsible for the object recognition and tracking separately.

More formally, given a video sequence $\mathcal{V} = [\textit{X}_{1}, \textit{X}_{2}, ..., \textit{X}_{T}]$ and the annotation mask of the first frame, we process the frames sequentially and maintain a memory bank to store the past frames and their predictions following~\cite{oh2019video,cheng2021stcn}. For the current frame $\textit{X}_{t} \in \mathcal{R}^{3 \times H \times W}$, we first extract $F_{res4}$ from ResNet~\cite{he2016deep} as our backbone features, which is shared by a pixel decoder to generate per-pixel embeddings and a Transformer decoder to inject localized features to learnable object queries in IS branch. While in the VOS branch, we apply an Enhanced Key Encoder to project backbone feature $F_{res4}$ to query key, which are compared with the memory key to perform semantic matching. Finally, the memory readout as well as multi-scale features from both backbone and pixel decoder are input to the VOS decoder to produce the final mask prediction. The architecture of \system is illustrated in Figure~\ref{fig:network}. Below, we first introduce IS and VOS branch in Sec.~\ref{subsec:is_branch} and Sec.~\ref{subsec:vos_branch}, respectively, and then elaborate how to obtain the final mask prediction in Sec~\ref{subsec:decoders}.

\subsection{Instance Segmentation Branch}
\label{subsec:is_branch}
As is described above, existing memory-based VOS models typically perform dense matching between the features of current frame and memory bank without mining the instance information, which therefore suffers from false matches when distinct object deformation or appearance changes happen. To address this issue, we explore to acquire the instance understanding capability from an instance segmentation (IS) branch, which is built upon an auxiliary query-based instance segmentation model~\cite{cheng2022masked}. Specifically, our IS branch consists of the following components: 

\vspace{0.02in}
\noindent \textbf{Pixel Decoder} takes $F_{res4}$ as input and generates per-pixel embeddings $F_{pixel} \in \mathbb{R}^{C_{\varepsilon} \times H/4 \times W/4}$ with alternate convolutional and upsampling layers, where $C_{\varepsilon}$ is the embedding dimension. In addition, we also input the feature pyramid $\{P_{i}\}_{i=0}^{2}$ produced by the pixel decoder with a resolution 1/32, 1/16 and 1/8 of the original image into both the transformer decoder and the VOS decoder, so as to fully take advantage of high-resolution instance features. For each resolution, we add both a sinusoidal positional embedding and a learnable scale-level embedding following~\cite{zhu2021deformable,cheng2022masked}.

\vspace{0.02in}
\noindent \textbf{Transformer Decoder} gathers the local information in features obtained by pixel decoder to a set of learnable object queries $q_{ins} \in \mathcal{R}^{N \times C_{d}}$ through masked attention~\cite{cheng2022masked}, where $N$ is a pre-defined hyper-parameter to indicate the number of object queries (which is set to 100, empirically), and $C_{d}$ is the query feature dimension. The masked attention can be formulated as:
\begin{equation}
    q_{l} = \mathrm{softmax}(\mathcal{M}_{l-1} + q_{l}k_{l}^{\mathrm{T}})v_{l} + q_{l-1},
\end{equation}
where $k_{l}$ and $v_{l}$ are the key and value embeddings projected from one resolution of $\{P_{i}\}_{i=0}^{2}$ (with $i$ corresponds to $l$), respectively. We add an auxiliary loss~\cite{cheng2022masked} to every intermediate Transformer decoder layer and $\mathcal{M}_{l-1}$ is the binarized (threshold = 0.5) mask prediction from the previous $(l-1)_{th}$ layer. Note that $q_{0}$ is initialized with $q_{ins}$ and finally updated to $\tilde{q}_{ins}$ as the final object queries.

\subsection{Video Object Segmentation Branch}
\label{subsec:vos_branch}
With the instance-specific information from the IS branch, our VOS branch further performs instance-augmented matching between the current frame and the memory bank in the feature space, so as to leverage long-range context information for mask generation. 

\vspace{0.02in}
\noindent \textbf{Enhanced Key Encoder} takes the updated object queries $\tilde{q}_{ins}$ and the backbone feature $F_{res4}$ as inputs to generate the query key of current frame with $C_{k}$-dimension. Specifically, we follow~\cite{cheng2021stcn} to first apply a $3\times3$ convolutional layer on top of $F_{res4}$ to obtain $Q_{g} \in \mathcal{R}^{C_{h} \times H/16 \times W/16}$, $C_{h}$ denotes the hidden dimension and we set it to the query feature dimension. Next, we aggregate the image features in $Q_{g}$ to $\tilde{q}_{ins}$ through a deformable attention layer~\cite{zhu2021deformable}:
\begin{equation}
    \tilde{q}_{vos} = \mathrm{DeformAttn}(\tilde{q}_{ins}, p, Q_{g}),
    \label{formulation:qe}
\end{equation}
where $p$ is a set of 2-d reference points. 

After that, we inject instance information in $\tilde{q}_{vos}$ to $Q_{g}$ reversely through a dot product (flattening operation is omitted for brevity), which is then activated by a sigmoid function and concatenated with $Q_{g}$ to get $Q_{cat}$. Finally, we apply a convolutional projection head on $Q_{cat}$ and further flatten it to the instance-aware query key $Q \in \mathcal{R}^{C_{k} \times H_{m} W_{m}}$, where $H_{m} = H / 16$ and $W_{m} = W / 16$. 

\vspace{0.02in}
\noindent \textbf{Memory Reading} first retrieves the memory key $K \in \mathcal{R}^{C_{K} \times T H_{m} W_{m}}$ and memory value $V \in \mathcal{R}^{C_{v} \times T H_{m} W_{m}}$ from memory bank, where $T$ is the current memory size, and $C_{v}$ denotes the value feature dimension. Then the similarity between $K$ and the above query key $Q$ is measured by calculating the affinity matrix:
\begin{equation}
   A_{i,j} = \frac{\mathrm{exp}(\mathrm{d}(K_{i}, Q_{j}))}{\sum_{i}(\mathrm{exp}(\mathrm{d}(K_{i}, Q_{j})))},
\end{equation}
where the subscript indexes the spatial location of query and memory key, and the distance function here we use is L2 distance following~\cite{cheng2021stcn,cheng2022xmem}. Note that we normalize the distance value by $\sqrt{C_{k}}$ as in~\cite{oh2019video,cheng2021stcn}.

With the affinity matrix $A$, the memory value $V$ could be aggregated through a weighted summation to obtain the readout features $F_{mem} \in \mathcal{R}^{C_{v} \times H_{m} W_{m}}$. Finally, we pass $F_{mem}$ to the VOS decoder for mask generation, which will be clarified later.

\vspace{0.02in}
\noindent \textbf{Memory Update} is executed once the prediction of the current frame is generated  during training and at a fixed interval during inference, which stores the memory key and memory value of current frame into the memory bank. We follow~\cite{cheng2021stcn,cheng2022xmem} to simply share the key between query and memory, \ie, the query key will be saved in the memory bank as a memory key if the current frame should be ``memorized''. While for the memory value, we first input the predicted mask to a lightweight backbone (ResNet18~\cite{he2016deep} is adopted in this paper), the last layer feature of which is concatenated with $F_{res4}$ to obtain $\tilde{V}_{cur}$ following~\cite{cheng2021stcn,cheng2022xmem}. Next, we further input $\tilde{V}_{cur}$ to two ResBlocks and a CBAM block~\cite{woo2018cbam} sequentially to get the memory value $V_{cur} \in \mathcal{R}^{C_{v} \times H_{m} W_{m}}$. In this paper, we describe the forward process for a single target object for readability. In the case of multi-object segmentation, an extra dimension is needed for $V$ to indicate the number of objects~\cite{cheng2021stcn}.

\begin{table*}[t]
\centering
  \setlength{\tabcolsep}{0pt}
  \begin{tabular*}{\linewidth}{@{\extracolsep{\fill}}lcc | cccc | cccc | ccccc@{}}
    \toprule
    \multirow{2}*{\textbf{Method}} & \multirow{2}*{\textbf{w/ BL30K}} && \multicolumn{3}{c}{\textbf{DAVIS16 validation}} && \multicolumn{3}{c}{\textbf{DAVIS17 validation}} && \multicolumn{5}{c}{\textbf{YT2018 validation}}  \\
    ~ & ~ && $\mathcal{J\&F}$ & $\mathcal{J}$ & $\mathcal{F}$ && $\mathcal{J\&F}$ & $\mathcal{J}$ & $\mathcal{F}$ && $\mathcal{G}$ & $\mathcal{J}_{s}$ & $\mathcal{F}_{s}$ & $\mathcal{J}_{u}$ & $\mathcal{F}_{u}$ \\
    \midrule
    STM~\cite{oh2019video} & \XSolidBrush && 89.3 & 88.7 & 89.9 && 81.8 & 79.2 & 84.3 &&  79.4 & 79.7 & 84.2 & 72.8 & 80.9 \\
    HMMN~\cite{shi2015hierarchical} & \XSolidBrush && 90.8 & 89.6 & 92.0 && 84.7 & 81.9 & 87.5 && 82.6 & 82.1 & 87.0 & 76.8 & 84.6 \\
    RPCM~\cite{xu2022reliable} & \XSolidBrush && 90.6 & 87.1 & 91.1 && 83.7 & 81.3 & 86.0 && 84.0 & 83.1 & 87.7 & 78.5 & 86.7 \\
    STCN~\cite{cheng2021stcn} & \XSolidBrush && 91.6 & 90.8 & 92.5 && 85.4 & 82.2 & 88.6 && 83.0 & 81.9 & 86.5 & 77.9 & 85.7 \\
    AOT~\cite{yang2021associating} & \XSolidBrush && 91.1 & 90.1 & 92.1 && 84.9 & 82.3 & 87.5 && 85.5 & 84.5 & 89.5 & 79.6 & 88.2 \\
    RDE~\cite{li2022recurrent} & \XSolidBrush && 91.1 & 89.7 & 92.5 && 84.2 & 80.8 & 87.5 && - & - & - & - & - \\
    XMem~\cite{cheng2022xmem} & \XSolidBrush && 91.5 & 90.4 & 92.7 && 86.2 & 82.9 & 89.5 && 85.7 & 84.6 & 89.3 & 80.2 & 88.7 \\
    DeAOT~\cite{yang2022deaot} & \XSolidBrush && 92.3 & 90.5 & 94.0 && 85.2 & 82.2 & 88.2 && 86.0 & 84.9 & 89.9 & 80.4 & 88.7 \\
    Ours & \XSolidBrush && \textbf{92.6} & \textbf{91.5} & \textbf{93.7} && \textbf{87.1} & \textbf{83.7} & \textbf{90.5} && \textbf{86.3} & \textbf{85.5} & \textbf{90.2} & \textbf{80.5} & \textbf{88.8} \\
    \midrule
    MiVOS~\cite{cheng2021mivos} & \Checkmark && 91.0 & 89.6 & 92.4 && 84.5 & 81.7 & 87.4 && 82.6 & 81.1 & 85.6 & 77.7 & 86.2 \\
    STCN~\cite{cheng2021stcn} & \Checkmark && 91.7 & 90.4 & 93.0 && 85.3 & 82.0 & 88.6 && 84.3 & 83.2 & 87.9 & 79.0 & 87.3 \\
    RDE~\cite{li2022recurrent} & \Checkmark && 91.6 & 90.0 & 93.2 && 86.1 & 82.1 & 90.0 && - & - & - & - & - \\
    XMem~\cite{cheng2022xmem} & \Checkmark && 92.0 & 90.7 & 93.2 && 87.7 & 84.0 & 91.4 && 86.1 & 85.1 & 89.8 & 80.3 & \textbf{89.2} \\
    Ours & \Checkmark && \textbf{92.8} & \textbf{91.8} & \textbf{93.8} && \textbf{88.2} & \textbf{84.5} & \textbf{91.9} && \textbf{86.7} & \textbf{86.1} & \textbf{90.8} & \textbf{81.0} & \textbf{89.0}  \\
    \bottomrule
  \end{tabular*}
  \vspace{-0.1in}
 \caption{Quantitative comparisons on the DAVIS 2016 val, DAVIS 2017 val, and YouTube-VOS 2018 val split. }
\label{tab:short}
\end{table*}

\subsection{Mask Prediction}
\label{subsec:decoders}
On top of the IS branch and VOS branch, we apply an auxiliary instance segmentation decoder and a video object segmentation decoder to generate the instance mask and video object mask predictions, respectively. Note that \textit{ the auxiliary instance segmentation decoder along with the pixel embedding will only be used during training and discarded during inference.}

\noindent \textbf{Instance Segmentation Decoder} inputs the updated object queries $\tilde{q}_{ins}$ to a linear classifier and a softmax activation function successively to yield category probabilities. Besides, a Multi-Layer Perceptron (MLP) with 2 hidden layers transforms $\tilde{q}_{ins}$ to the corresponding mask embeddings. Finally, we obtain each binary mask prediction $\hat{M}_{ins}$ via a dot product between the mask embedding and per-pixel embeddings $F_{pixel}$.

\begin{figure}[t]
  \centering
  \includegraphics[width=\linewidth]{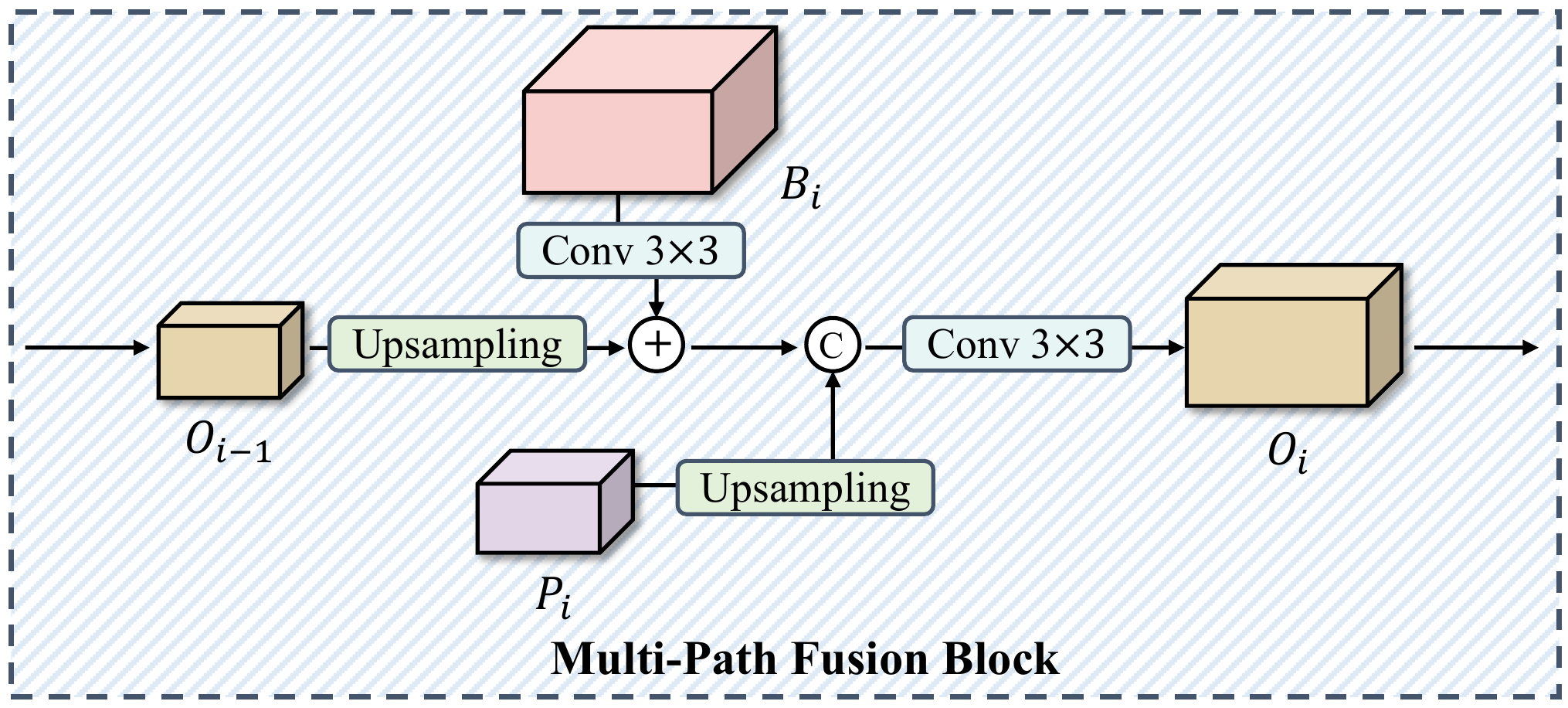}
   \vspace{-0.25in}
   \caption{Illustration of the Multi-Path Fusion (MPF) block.}
   \label{fig:vos_decoder}
\end{figure}

\vspace{0.02in}
\noindent \textbf{Video Object Segmentation Decoder} fuses the memory readout $F_{mem}$, multi-scale features from backbone $\{B_{i}\}_{i=0}^{2}$\footnote{We follow the previous work~\cite{oh2019video,cheng2021stcn,cheng2022xmem} to adopt the features with stride = 4, 8, and 32. } and pixel decoder $\{P_{i}\}_{i=0}^{2}$ with a multi-path fusion (MPF) block to make use of the fine-grained details reserved in high-resolution instance-aware features. The MPF block could be formulated as:
\begin{equation}
    O_{i} = \mathrm{MPF}(O_{i-1}, B_{i}, P_{i}),
    \label{formulation:mpf}
\end{equation}
$O_{i-1}$ is output by the previous MPF block, which is initialized with $F_{mem}$. Specifically, we first input $B_{i}$ and $P_{i}$ to $3\times3$ convolutional layers to align their feature dimensions with $O_{i-1}$ to obtain $\tilde{B}_{i}$ and $\tilde{P}_{i}$, separately. Next, we concatenate the sum of upsampled $O_{i-1}$ and $\tilde{B}_{i}$ with the upsampled $\tilde{P}_{i}$, the result of which is finally input to a residual block to get $O_{i}$. The detailed architecture of MPF block is illustrated in Figure~\ref{fig:vos_decoder}. Note that the final layer of the decoder produces a mask with stride = 4, and we bilinearly upsample it to the original resolution. 

\section{Experiments}
\label{sec:exp}
\subsection{Implementation Details}
\label{subsec:implementation}
\noindent \textbf{Training.} The instance segmentation (IS) branch and video object segmentation (VOS) branch are jointly trained with different supervisory signals. We train the IS branch on a standard instance segmentation dataset COCO~\cite{lin2014microsoft} with a resolution of 384 (compatible with VOS branch). While for the VOS branch, we follow~\cite{oh2019video,seong2020kernelized,liang2020video,cheng2021stcn,cheng2022xmem} to firstly pretrain our network on deformed static images~\cite{cheng2020cascadepsp,li2020fss,shi2015hierarchical,wang2017learning,zeng2019towards}, and then perform the main training on YouTube-VOS~\cite{xu2018YouTube}
and DAVIS~\cite{pont20172017}. Note that we also pretrain on BL30K~\cite{chang2015shapenet,denninger2019blenderproc,cheng2021mivos} optionally to further boost the performance following~\cite{cheng2021stcn,cheng2022xmem}, and the models pretrained on additional data are denoted with an asterisk $(*)$.  

The IS branch is supervised with a combination of mask loss and classification loss, where the mask loss consists of weighted binary cross-entropy loss and dice loss~\cite{milletari2016v}. The VOS branch is supervised with bootstrapped cross entropy loss and dice loss following~\cite{yang2021associating}. The static image pretraining lasts 150K iterations with a batch size of 56 and a learning rate of 4e-5. While the main training lasts 110K iterations with a batch size 16 and a learning rate 2e-5. The complete model is optimized with AdamW~\cite{kingma2014adam,loshchilov2017decoupled} with a weight decay of 0.05.  We load the COCO-pretrained Mask2Former~\cite{cheng2022masked} to initialize our instance segmentation branch, and use 0.1$\times$ learning rate for these parameters. The overall learning rate is decayed by a factor of 10 after the first 80K iterations.

\begin{figure*}[t]
  \centering
   \includegraphics[width=\linewidth]{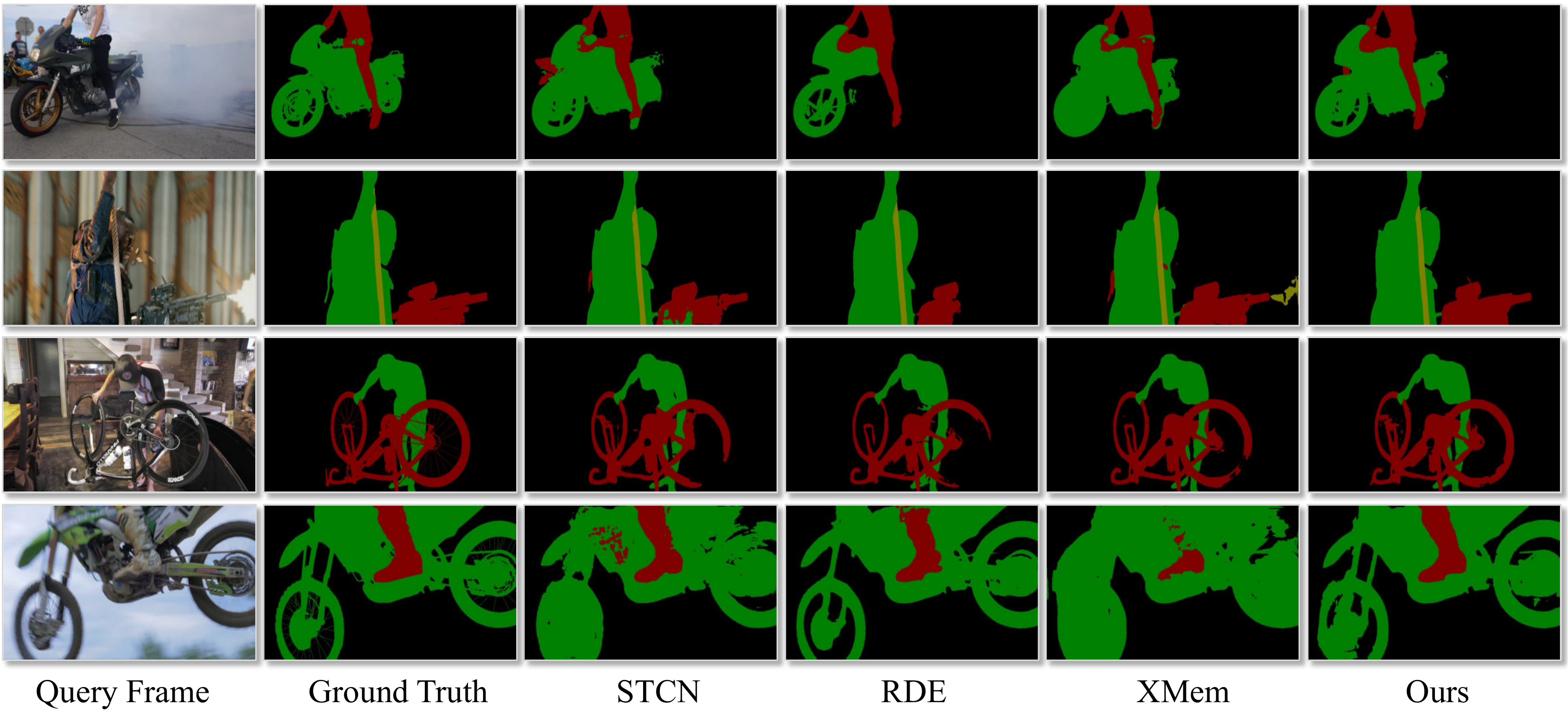}
   \vspace{-0.25in}
   \caption{Qualitative comparisons between \system and several state-of-the-art memory-based VOS models, including RDE~\cite{li2022recurrent}, STCN~\cite{cheng2021stcn}, and XMem~\cite{cheng2022xmem}.}
   \label{fig:comparison}
\end{figure*}

\vspace{0.02in}
\noindent \textbf{Inference.} We follow previous work~\cite{oh2019video,seong2020kernelized,cheng2021stcn,cheng2022xmem} to memorize every 5th frame during inference. Specially, we implement the memory as a first-in-first-out (FIFO) queue, and restrict the maximum memory size to 16 to improve the inference speed. Note that the first frame and its mask are always reserved to provide accurate reference information. We adopt Top-K filter~\cite{cheng2021mivos,cheng2021stcn,cheng2022xmem} for memory-reading augmentation, with K set to 20.

\vspace{0.02in}
\noindent \textbf{Evaluation datasets and metrics.} We evaluate the performance of \system on standard VOS datasets DAVIS~\cite{perazzi2016benchmark,pont20172017} and YouTube-VOS~\cite{xu2018YouTube}. DAVIS 2016~\cite{perazzi2016benchmark} is a single-object VOS benchmark and DAVIS 2017~\cite{pont20172017} extends it to a multi-object version. YouTube-VOS~\cite{xu2018YouTube} is the large-scale benchmark for multi-object VOS, which also includes unseen categories in the validation set to measure the generalization ability. We report the results on 474 and 507 validation videos in the 2018 and 2019 versions (denoted as ``YT2018" and ``YT2019" in the following Tables respectively). We use mean Jaccard $\mathcal{J}$ index and mean boundary $\mathcal{F}$ score, along with mean $\mathcal{J\&F}$ to evaluate segmentation accuracy. Note that for YouTube-VOS, we report the results on both seen and unseen categories, along with the averaged overall score $\mathcal{G}$. 

\subsection{Comparison with State-of-the-art Methods}
\begin{table}[t]
\centering
  \setlength{\tabcolsep}{0pt}
  \begin{tabular*}{\linewidth}{@{\extracolsep{\fill}}lc | cccc | ccccc@{}}
    \toprule
    \multirow{2}*{\textbf{Method}} && \multicolumn{3}{c}{\textbf{D17 test-dev}}  && \multicolumn{5}{c}{\textbf{YT2019 validation}}  \\
    ~ && $\mathcal{J\&F}$ & $\mathcal{J}$ & $\mathcal{F}$ && $\mathcal{G}$ & $\mathcal{J}_{s}$ & $\mathcal{F}_{s}$ & $\mathcal{J}_{u}$ & $\mathcal{F}_{u}$ \\
    \midrule
    HMMN~\cite{seong2021hierarchical} && 78.6 & 74.7 & 82.5 && 82.5 & 81.7 & 86.1 & 77.3 & 85.0 \\
    STCN~\cite{cheng2021stcn} && 76.1 & 73.1 & 80.0 && 82.7 & 81.1 & 85.4 & 78.2 & 85.9 \\
    RPCM~\cite{xu2022reliable} && 79.2 & 75.8 & 82.6 && 83.9 & 82.6 & 86.9 & 79.1 & 87.1 \\
    AOT~\cite{yang2021associating} && 79.6 & 75.9 & 83.3 &&  85.3 & 83.9 & 88.8 & 79.9 & 88.5 \\
    RDE~\cite{li2022recurrent} && 77.4 & 73.6 & 81.2 && 81.9 & 81.1 & 85.5 & 76.2 & 84.8 \\
    XMem~\cite{cheng2022xmem} && 81.0 & 77.4 & 84.5 && 85.5 & 84.3 & 88.6 & 80.3 & 88.6 \\
    DeAOT~\cite{yang2022deaot} && 80.7 & 76.9 & 84.5 && 85.9 & 84.6 & 89.4 & 80.8 & 88.9 \\
    Ours && \textbf{82.8} & \textbf{79.3} & \textbf{86.2} && \textbf{86.1} & \textbf{85.2} & \textbf{89.7} & \textbf{80.7} & \textbf{88.9} \\
    \midrule
    MiVOS$^{*}$~\cite{cheng2021mivos} && 78.6 & 74.9 & 82.2 && 82.4 & 80.6 & 84.7 & 78.1 & 86.4  \\
    STCN$^{*}$~\cite{cheng2021stcn} && 77.8 & 74.3 & 81.3 && 84.2 & 82.6 & 87.0 & 79.4 & 87.7 \\
    RDE~\cite{li2022recurrent} && 78.9 & 74.9 & 82.9 && 83.3 & 81.9 & 86.3 & 78.0 &  86.9\\
    XMem$^{*}$~\cite{cheng2022xmem} && 81.2 & 77.6 & 84.7 && 85.8 & 84.8 & 89.2 & 80.3 & 88.8 \\
    Ours$^{*}$ && \textbf{84.0} & \textbf{80.1} & \textbf{87.8} && \textbf{86.3} & \textbf{85.2} & \textbf{89.7} & \textbf{81.0} & \textbf{89.1} \\
    \bottomrule
  \end{tabular*}
  \vspace{-0.1in}
 \caption{Results on DAVIS 2017 (D17) test-dev and YouTube-VOS 2019 validation. $^{*}$ denotes BL30K is adopted for pretraining. }
 \vspace{-0.1in}
\label{tab:addition}
\end{table}

The comparison results between \system and existing state-of-the-art VOS models on DAVIS 2016 validation, DAVIS 2017 validation, and YouTube-VOS 2018 validation are listed in Table~\ref{tab:short}. We can see that without incorporating BL30K as addition training data, our method achieves top-ranked performance on both single-object and multi-object VOS benchmarks, \ie, 92.6\%, 87.1\%, 86.3\% in terms of $\mathcal{J\&F}$ on DAVIS 2016 $\&$ 2017, YouTube-VOS 2018 validation split, respectively, even surpassing existing methods that are pretrained on BL30K. Adopting BL30K as additional training data can further 
boost the performance of \system. We also report the results on DAVIS 2017 test-dev and YouTube-VOS 2019 validation split in Table~\ref{tab:addition}, and \system also outperforms all the baseline methods. Even though our method adopts a simpler memory mechanism than existing methods like~\cite{seong2021hierarchical,xu2022reliable,cheng2022xmem}, we still achieve superior performance. This highlights that introducing instance understanding to conduct instance-augmented matching is super helpful and clearly outperforms the vanilla semantic matching used in the existing methods~\cite{oh2019video,cheng2021stcn,cheng2022xmem}. 

We further visualize the segmentation results of \system on some representative challenging cases with dramatic movements (\eg, shooting and motor cross-jump), and compare with state-of-the-art memory-based VOS models including STCN~\cite{cheng2021stcn}, RDE~\cite{li2022recurrent}, and XMem~\cite{cheng2022xmem} in Figure~\ref{fig:comparison}. We can see that RDE~\cite{li2022recurrent} and STCN~\cite{cheng2021stcn} struggle with occlusions incurred by smoke and confusing objects, respectively. XMem~\cite{cheng2022xmem} produces more competitive results, which however fails to generate sharp boundaries for the motorcycle rims. Our method, by contrast, generates more accurate and clear masks on these challenging cases. This suggests that the instance-aware representations learned from the instance segmentation branch could facilitate our model to derive instance-discriminative features.

\subsection{Discussion}
\noindent \textbf{Impact of query enhancement and MPF block.} The query enhancement (QE) (Equation~\ref{formulation:qe}) is used to inject instance-specific into query key, while the multi-path fusion (MPF) block (Equation~\ref{formulation:mpf}) is designed to incorporate high-resolution instance-aware features for fine-grained detail prediction. To evaluate their effectiveness, we remove QE and MPF separately from \system and evaluate the segmentation performance on DAVIS 2017 validation (DAVIS17 val) and YouTube-VOS 2018 validation (YT2018 val) split.

\begin{table}[!ht]
\centering
  \setlength{\tabcolsep}{0pt}
  \begin{tabular*}{\linewidth}{@{\extracolsep{\fill}}lc | cccc | ccccc @{}}
    \toprule
    \multirow{2}{0.6in}{\textbf{Method}} && \multicolumn{3}{c}{\textbf{DAVIS17 val}} && \multicolumn{5}{c}{\textbf{YT2018 val}}  \\
    ~ && $\mathcal{J\&F}$ & $\mathcal{J}$ & $\mathcal{F}$ &&  $\mathcal{G}$ & $\mathcal{J}_{s}$ & $\mathcal{F}_{s}$ & $\mathcal{J}_{u}$ & $\mathcal{F}_{u}$ \\
    \midrule
    w/o QE $\&$ MPF && 85.3 & 82.0 & 88.6 && 83.0 & 84.0 & 75.7 & 88.5 & 83.8 \\
    w/o QE && 85.7	& 82.4 & 88.9 && 84.4 & 85.1 & 77.4 & 89.8 & 85.5 \\
    w/o MPF && 86.2 & 83.0 & 89.5 && 85.6 & 85.0 & 90.4 & 79.4 & 87.5 \\
    Ours && \textbf{87.1} & \textbf{83.7} & \textbf{90.5} && \textbf{86.3} & \textbf{85.5} & \textbf{90.2} & \textbf{80.5} & \textbf{88.8} \\
    \bottomrule
  \end{tabular*}
  \vspace{-0.1in}
 \caption{Results on DAVIS 2017 validation and YouTube-VOS validation split w/ and w/o query enhancement (QE) and multi-path fusion (MPF) block. }
\label{tab:components}
\end{table}

The quantitative results are listed in Table~\ref{tab:components}. We can observe that without QE, the $\mathcal{J\&F}$ value decreases by 1.4\% on DAVIS17 val and 1.9\% on YT2018 val, while without MPF, the $\mathcal{J\&F}$ value decreases by 0.9\% and 0.7\%, respectively. The performance degradation validates that the use of the above components both effectively improves the performance of our model.

\vspace{0.02in}
\noindent \textbf{Impact of weight initialization and joint training for the IS branch.} The IS branch is built upon a instance segmentation model~\cite{cheng2022masked} to acquire instance-specific information. In our implementation, we load the weights from Mask2Former~\cite{cheng2022masked} pretrained on COCO~\cite{lin2014microsoft} and perform joint training on both IS task and VOS task to prevent the catastrophic forgetting. To study the effect of weight initialization and joint training, we conduct experiments under different settings and compare the results in Table~\ref{tab:instance_segmentation}.

\begin{table}[!ht]
\centering
  \setlength{\tabcolsep}{0pt}
  \begin{tabular*}{\linewidth}{@{\extracolsep{\fill}}ccc | cccc | ccccc @{}}
    \toprule
    \multirow{2}{0.25in}{\textbf{LW}} & \multirow{2}{0.25in}{\textbf{JT}} && \multicolumn{3}{c}{\textbf{DAVIS17 val}} && \multicolumn{5}{c}{\textbf{YT2018 val}}  \\
    ~ & ~ && $\mathcal{J\&F}$ & $\mathcal{J}$ & $\mathcal{F}$ &&  $\mathcal{G}$ & $\mathcal{J}_{s}$ & $\mathcal{F}_{s}$ & $\mathcal{J}_{u}$ & $\mathcal{F}_{u}$ \\
    \midrule
    \XSolidBrush & \XSolidBrush && 78.6 & 75.6 & 81.5 && 78.6 & 79.2 & 83.5 & 72.3 & 79.4 \\
    \XSolidBrush & \Checkmark && 80.0 & 76.9 & 83.1 && 80.6 & 80.1 & 84.5 & 74.5 & 83.2\\ 
    \Checkmark & \XSolidBrush && 82.0 & 79.2 & 84.4 && 81.3 & 80.7 & 85.4 & 75.5 & 83.6 \\
    \Checkmark & \Checkmark && \textbf{87.1} & \textbf{83.7} & \textbf{90.5} && \textbf{86.3} & \textbf{85.5} & \textbf{90.2} & \textbf{80.5} & \textbf{88.8} \\
    \bottomrule
  \end{tabular*}
  \vspace{-0.1in}
 \caption{Results on DAVIS 2017 validation and YouTube-VOS validation split w/ and w/o loading the weights from pretrained Mask2Former~\cite{cheng2022masked} (\textbf{LW}) and joint training (\textbf{JT}). }
\label{tab:instance_segmentation}
\end{table}

\begin{figure}[!ht]
  \centering
   \includegraphics[width=\linewidth]{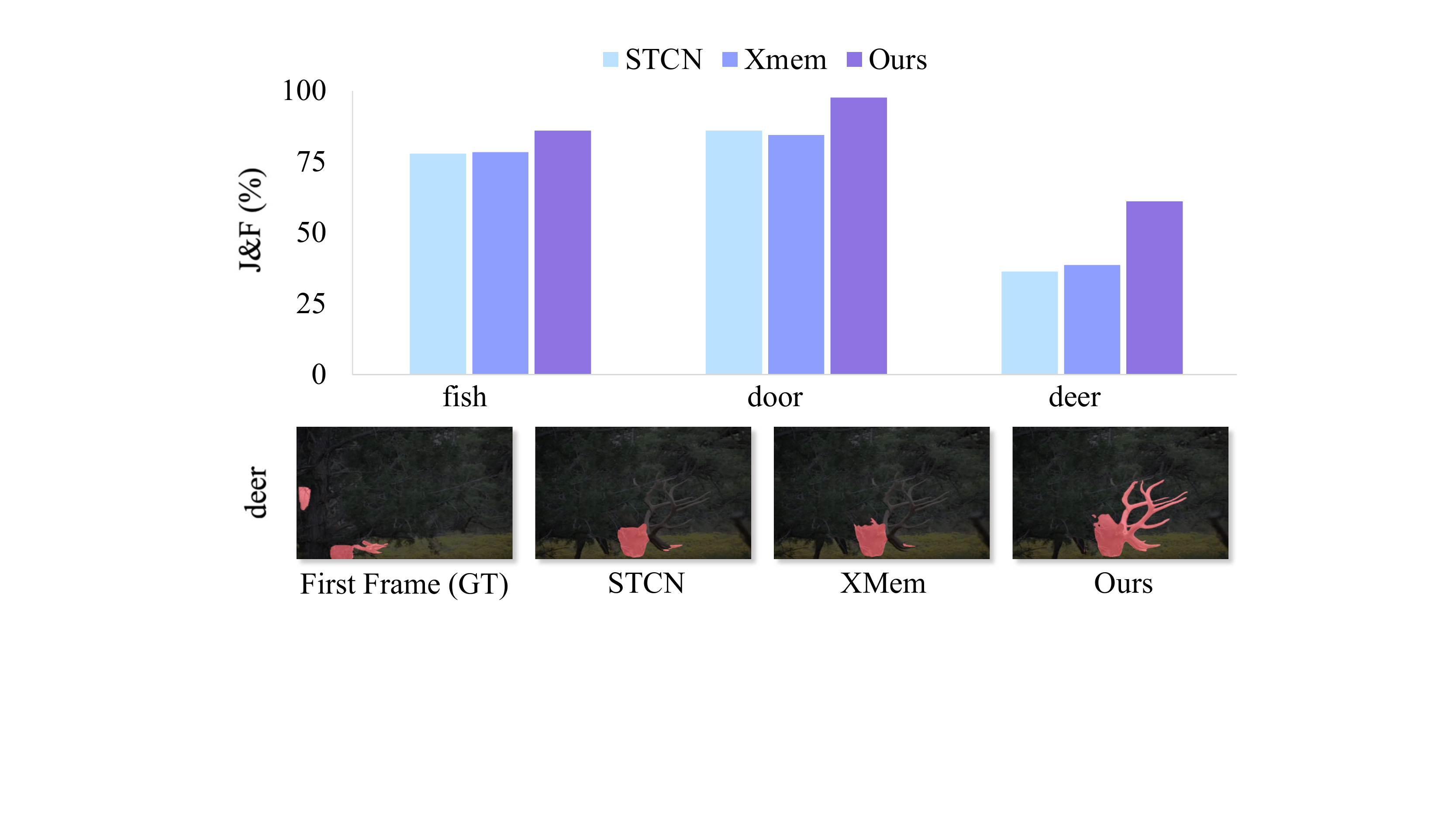}
   \vspace{-0.15in}
   \caption{$J\&F$ metric of STCN~\cite{cheng2021stcn}, XMem~\cite{cheng2022xmem}, and \system on three videos whose categories do not appear in the COCO dataset.}
   \label{fig:openset}
   \vspace{-0.1in}
\end{figure}

The drastic performance drop in the first row indicates that instance-aware representations are critical to the generation of accurate masks. Performing joint training from scratch brings about minor improvements, but it is difficult for the VOS branch to learn useful instance information from IS branch in the beginning. Initializing the weights from Mask2Former~\cite{cheng2022masked} improves the performance more significantly, which however, would gradually lose the instance segmentation capability without joint training. In contrast, the combined weight initialization from Mask2Former and joint training achieves the best results. We would also like to point out while we use Mask2Former for initialization, the IS branch can be easily replaced with any query-based instance segmentation model.

In addition, considering that VOS is essentially a category-agnostic task but the IS branch is trained on a close set, we further show the performance of \system on objects that do not appear in COCO, \eg, fish, door, and deer\footnote{These objects correspond to the f78b3f5f34, 4d6cd98941, and f6ed698261 video in YouTube-VOS 2018 val.}, and compare with STCN and XMem in Figure~\ref{fig:openset}. The quantitative comparison is also displayed. We can see that our method still performs well on these objects and generates more accurate masks than existing methods. This indicates that joint training allows our method to develop generalizable instance differentiation capability even if the IS branch is trained on a close-set instance segmentation dataset.

\begin{figure}[!ht]
\begin{minipage}[t]{0.49\linewidth}
\centering
\includegraphics[width=1.65in]{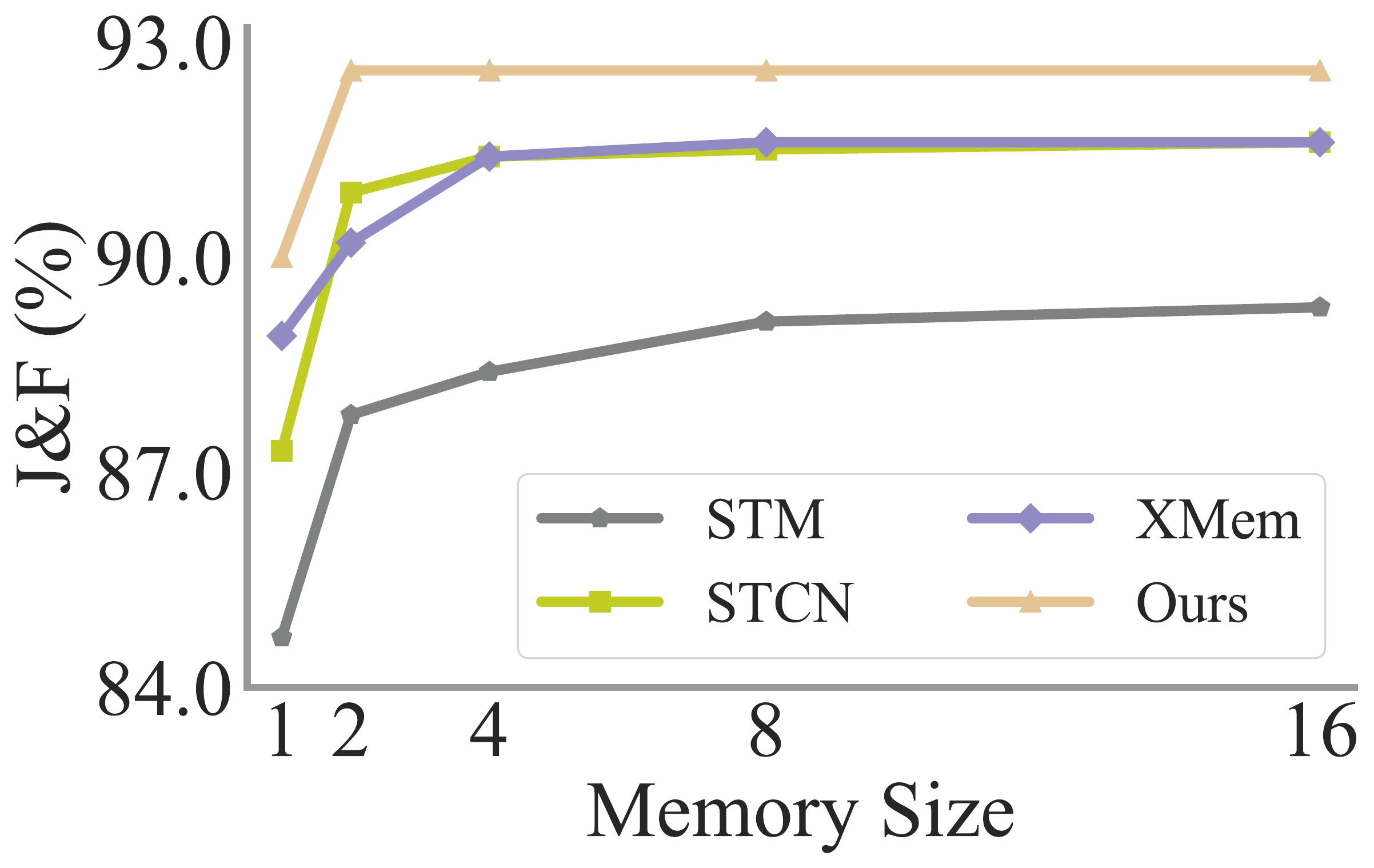}
\end{minipage}
\begin{minipage}[t]{0.49\linewidth}
\centering
\includegraphics[width=1.65in]{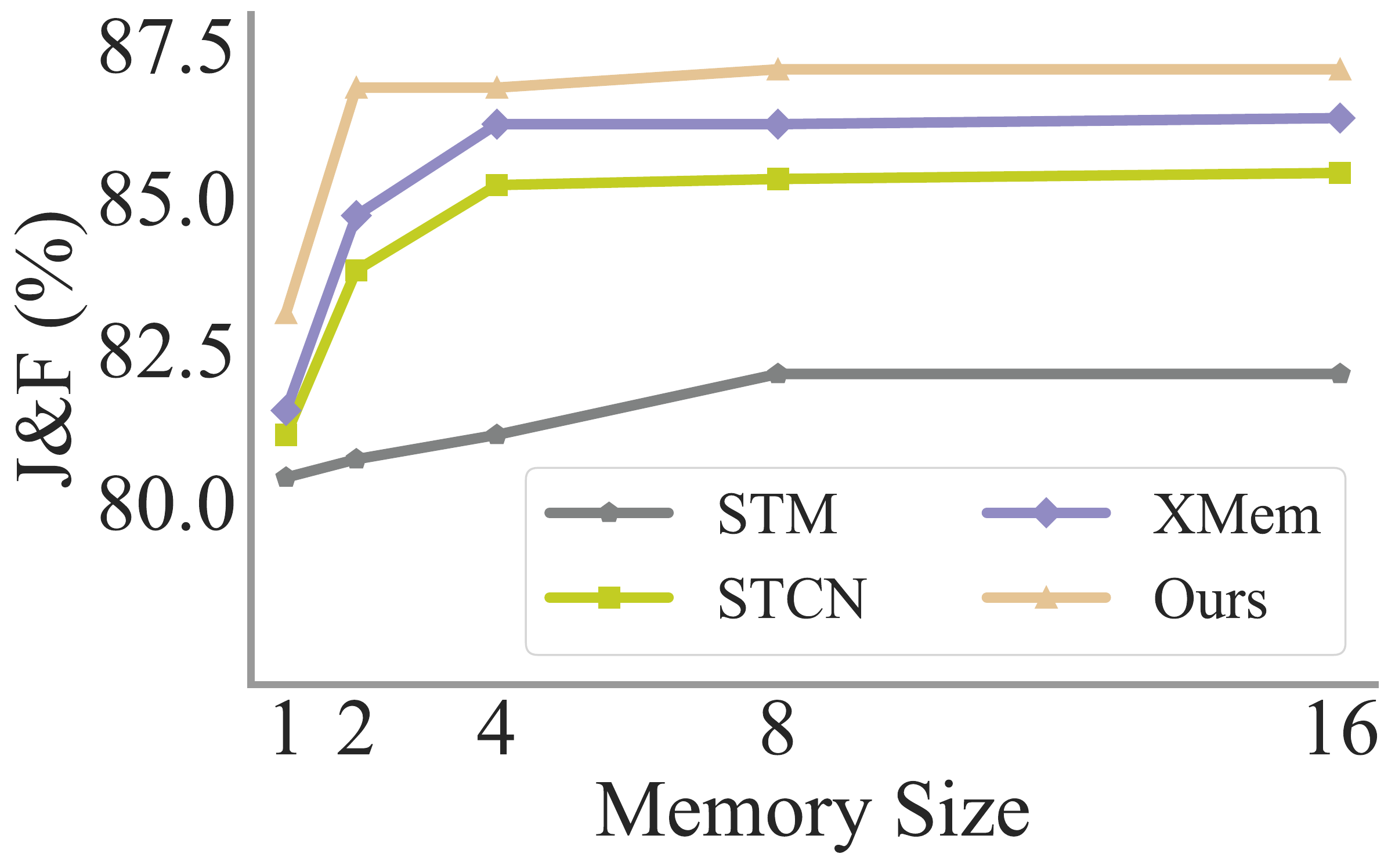}
\end{minipage}
\vspace{-0.1in}
\caption{Tradeoffs between the maximum memory size and $\mathcal{J\&F}$ on DAVIS 2016 (left) and 2017 (right) validation split.}
\label{fig:tradeoffs}
\end{figure}

\vspace{0.05in}
\noindent \textbf{Trade-off between memory size and segmentation performance.} As mentioned in Sec.~\ref{subsec:implementation}, we implement the memory bank as a first-in-first-out (FIFO) queue with a maximum size. To further investigate the behavior of \system, we dynamically adjust the maximum memory size and observe the trend of performance variation (\ie, $\mathcal{J\&F}$) on DAVIS 2016 $\&$ 2017 validation split. We also re-implement the memory bank of several existing memory-based models (including STM~\cite{oh2019video}, STCN~\cite{cheng2021stcn}, and XMem~\cite{cheng2022xmem}) as FIFO queues, and compare their results with \system in Figure~\ref{fig:tradeoffs}.

We can see that increasing the memory size always improves the segmentation performance for all these methods,  since more contextual information is used. It is noteworthy that, our method achieves competitive results by relying on a smaller memory size, \eg, when the memory size is set to 2, the $\mathcal{J\&F}$ value is only 0.3 away from the highest point for \system, but 1.6 for XMem on DAVIS 2017 validation split. This demonstrates the superiority of instance-augmented matching compared with vanilla semantic matching.

\vspace{0.05in}
\noindent \textbf{Results with different training data.} In the main experiment, we follow previous methods~\cite{oh2019video,cheng2021mivos,cheng2021stcn,cheng2022xmem} to first pretrain our model on static images (and BL30K optionally) for fair comparisons. To study the effects of pretraining on the final segmentation results, we additionally conduct experiments to train \system on DAVIS 2017~\cite{pont20172017} only, YouTube-VOS 2019~\cite{xu2018YouTube} only, and a mix of both. The comparison with existing models are shown in Table~\ref{tab:data}. We can see that \system achieves competitive results even without incorporating static images and BL30K for pretraining, outperforming all the baseline models by a large margin. When gradually increasing the scale of the training data, the performance of our method can be further boosted.

\begin{table}[!ht]
\centering
  \renewcommand\arraystretch{1.1}
  \setlength{\tabcolsep}{0pt}
  \begin{tabular*}{\linewidth}{@{\extracolsep{\fill}}lc | cccc | ccccc @{}}
    \toprule
    \multirow{2}{0.7in}{\textbf{Method}} && \multicolumn{3}{c}{\textbf{DAVIS17 val}} && \multicolumn{5}{c}{\textbf{YT2018 val}}  \\
    ~ && $\mathcal{J\&F}$ & $\mathcal{J}$ & $\mathcal{F}$ &&  $\mathcal{G}$ & $\mathcal{J}_{s}$ & $\mathcal{F}_{s}$ & $\mathcal{J}_{u}$ & $\mathcal{F}_{u}$ \\
    \midrule
    SST$\ddagger$~\cite{duke2021sstvos} && 82.5 & 79.9 & 85.1 && 81.7 & 81.2 & - & 76.0 & - \\
    CFBI+$\ddagger$~\cite{yang2021collaborative} && 82.9 & 80.1 & 85.7 && 82.0 & 81.2 & 86.0 & 76.2 & 84.6 \\
    JOINT$\ddagger$~\cite{mao2021joint} && 83.5 & 80.8 & 86.2 && 83.1 & 81.5 & 85.9 & 78.7 & 86.5 \\
    XMem$\ddagger$ && 84.5 & - & - && 84.3 & - & - & - & -\\
    Ours$\ddagger$ && 85.2 & 82.1 & 88.3 && 84.7 & 84.5 & 89.1 & 78.2 & 87.0 \\
    \cmidrule{1-1} \cmidrule{3-11}
    D only && 77.5 & 75.6 & 79.4 && - & - & - & - & - \\
    Y only && - & - & - && 84.9 & 84.0 & 88.8 & 78.8 &  88.0 \\
    S + D + Y && 87.1 & 83.7 & 90.5 && 86.3 & 85.5 & 90.2 & 80.5 & 88.8 \\
    S + D + B + Y && 88.2 & 84.5 & 91.9 && 86.7 & 86.1 & 90.8 & 81.0 & 89.0 \\
    \bottomrule
  \end{tabular*}
  \vspace{-0.1in}
 \caption{Results on DAVIS 2017 validation and YouTube-VOS validation split with different training data. D: DAVIS 2017, Y: YouTube 2019, S: static images, B: BL30K.  $\ddagger$ denotes pretraining on the combined DAVIS and YouTube-VOS data (\ie, D + Y). }
\label{tab:data}
\vspace{0.1in}
\end{table}

\vspace{0.05in}
\noindent \textbf{Multi-scale Inference.} Multi-scale evaluation is a commonly used trick in segmentation tasks~\cite{chandra2016fast,cheng2021stcn,cheng2022xmem} to boost the performance by merging the results of inputs under different data augmentations. Here we follow XMem~\cite{cheng2022xmem} to apply image scaling and vertical mirroring and simply
average the output probabilities to obtain the final masks. 

\begin{table}[!ht]
\centering
  \renewcommand\arraystretch{0.8}
  \setlength{\tabcolsep}{0pt} 
  \begin{tabular*}{\linewidth}{@{\extracolsep{\fill}}lcc | cccc | ccc @{}}
    \toprule
    \multirow{2}*{\textbf{Method}} & \multirow{2}*{\textbf{MS}} && \multicolumn{3}{c}{\textbf{D16 val}} && \multicolumn{3}{c}{\textbf{D17 val}} \\
    ~ & ~ && $\mathcal{J\&F}$ & $\mathcal{J}$ & $\mathcal{F}$ && $\mathcal{J\&F}$ & $\mathcal{J}$ & $\mathcal{F}$\\
    \midrule
    CFBI~\cite{yang2020collaborative} & \Checkmark && 90.7 & 89.6 & 91.7 && 83.3 & 80.5 & 86.0\\
    XMem~\cite{cheng2022xmem}  & \Checkmark && 92.7 & 92.0 & 93.5 && 88.2 & 85.4 & 91.0 \\
    Ours  & \XSolidBrush && 92.6 & 91.5 & 93.7 && 87.1 & 83.7 & 90.5 \\
    Ours  & \Checkmark && 92.9 & 92.2 & 93.6 && 88.6 & 85.8 & 91.4 \\
    Ours$^{*}$  & \XSolidBrush && 92.8 & 91.8 & 93.8 && 88.2 & 84.5 & 91.9 \\
    Ours$^{*}$  & \Checkmark && 93.4 & 92.5 & 94.2 && 89.8 & 86.7 & 93.0 \\
    \bottomrule
  \end{tabular*}
 \caption{Results on DAVIS 2017 validation and YouTube-VOS validation split with different training data. D: DAVIS 2017, Y: YouTube 2019, S: static images, B: BL30K.  $\ddagger$ denotes pretraining on the combined DAVIS and YouTube-VOS data. }
\label{tab:ms}
\end{table}

The results in Table~\ref{tab:ms} imply that multi-scale inference improves the performance of \system by 0.3\% and 1.7\% in terms of $\mathcal{J\&F}$ on DAVIS 2016 / 2017 validation split, and \system still outperforms existing methods.

\vspace{0.05in}
\noindent \textbf{Results on Long video datasets.} In order to further evaluate the long-term performance of \system, we additionally test our method on the Long-time Video dataset~\cite{liang2020video}, which contains three videos with more than 7,000 frames in total for validation. Considering the video duration is longer and the target object(s) will undergo distinct appearance deformation or scale variations, we set the maximum memory size to 64 during inference. The comparison results are shown in Table~\ref{tab:long}.

\begin{table}[!ht]
\centering
  \setlength{\tabcolsep}{0pt}
  \begin{tabular*}{0.9\linewidth}{@{\extracolsep{\fill}}l | ccc }
    \toprule
    \multirow{2}{1in}{\textbf{Method}} & \multicolumn{3}{c}{\textbf{Long-time Video}}  \\
    ~ & $\mathcal{J\&F}$ & $\mathcal{J}$ & $\mathcal{F}$ \\
    \midrule
    RMNet~\cite{xie2021efficient} & 59.8 & 59.7 & 60.0 \\
    JOINT~\cite{mao2021joint} & 67.1 & 64.5 & 69.6 \\
    STM~\cite{oh2019video} & 80.6 & 79.9 & 81.3 \\ 
    HMMN~\cite{shi2015hierarchical} & 81.5 & 79.9 & 83.0 \\
    STCN~\cite{cheng2021stcn} & 87.3 & 85.4 & 89.2  \\
    AOT~\cite{yang2021associating} & 84.3 & 83.2 & 85.4 \\
    AFB-URR~\cite{liang2020video} & 83.7 & 82.9 &  84.5 \\
    XMem~\cite{cheng2022xmem} & 89.8 & 88.0 & 91.6 \\
    Ours & \textbf{90.0} & \textbf{88.3} & \textbf{91.7} \\
    \bottomrule
  \end{tabular*}
 \caption{Results on the Long-time Video dataset~\cite{liang2020video}. }
\label{tab:long}
\end{table}

We can observe that \system again achieves the best segmentation results measured in different metrics. It is worth mentioning that \system beats the methods specifically designed for long videos, \eg, AFB-URR~\cite{liang2020video} and XMem~\cite{cheng2022xmem}. We believe the performance gain is resulted from taking advantage of the instance information in the query frame to facilitate the semantic matching.

\begin{figure}[!ht]
  \centering
   \includegraphics[width=\linewidth]{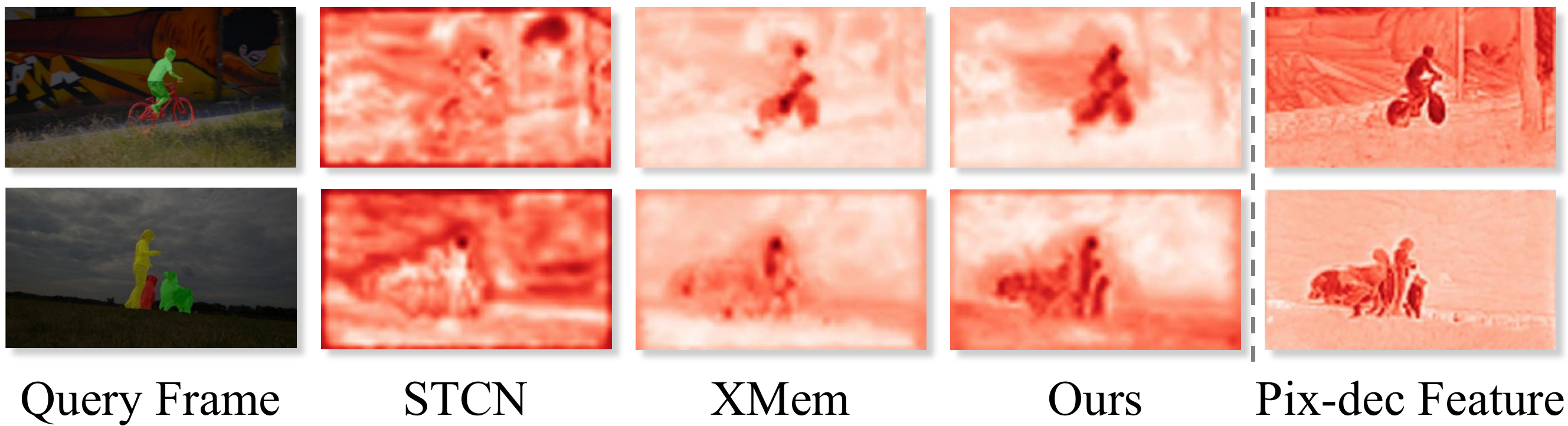}
   \vspace{-0.15in}
   \caption{Visualization of the query frame, memory readout features of STCN~\cite{cheng2021stcn}, XMem~\cite{cheng2022xmem}, our method, and the instance-aware features of the highest resolution $P_{2}$ from the pixel-decoder.}
   \label{fig:visualization}
\end{figure}

\vspace{0.05in}
\noindent \textbf{Visualization of Readout Features.} We visualize the readout features (\ie, $F_{mem}$ in Sec.~\ref{subsec:vos_branch}) of several memory-based VOS models and \system in Figure~\ref{fig:visualization} to further compare the vanilla semantic matching without instance understanding and instance-augmented matching. The high resolution feature $P_{2}$ from the pixel-decoder (Sec.~\ref{subsec:is_branch}) is also displayed. We can see that with the enhanced query key, the instance information in our readout features is more clear and distinguishable. In addition, the abundant details in high-resolution instance-aware features also help \system to produce sharp boundaries.

\section{Conclusion}
\label{sec:conclusion}
This paper proposes to incorporate instance understanding into memory-based matching for improved video object segmentation. To achieve this goal, a two-branch network \system is introduced, where the instance segmentation (IS) branch derives instance-aware representations of current frame and the video object segmentation (VOS) branch maintains a memory bank for spatial-temporal matching. We enhance the query key with the well-learned object queries from IS branch to inject the instance-specific information, with which the instance-augmented matching with memory bank is performed. Furthermore, we fuse the memory readout with multi-scale features from instance segmentation decoder through a carefully-designed multi-path fusion block. Extensive experiments conducted on both single-object and multi-object benchmarks demonstrate the effectiveness of the proposed method.

In addition to working towards more superior segmentation performance, another line of work~\cite{liang2020video,xie2021efficient,xu2022accelerating} also explore the efficient memory storage to improve the inference efficiency of memory-based methods. Therefore, \system can be combined with these approaches to develop both accurate and efficient VOS models.

{\small
\bibliographystyle{ieee_fullname}
\bibliography{egbib}
}
\appendix
\section{Multi-scale Inference.}
Multi-scale evaluation is a commonly used trick in segmentation tasks~\cite{chandra2016fast,cheng2021stcn,cheng2022xmem} to boost the performance by merging the results of inputs under different data augmentations. Here we follow XMem~\cite{cheng2022xmem} to apply image scaling and vertical mirroring and simply
average the output probabilities to obtain the final masks. 

\begin{table}[!ht]
	\centering
	\renewcommand\arraystretch{0.8}
	\setlength{\tabcolsep}{0pt} 
	\begin{tabular*}{\linewidth}{@{\extracolsep{\fill}}lcc | cccc | ccc @{}}
		\toprule
		\multirow{2}*{\textbf{Method}} & \multirow{2}*{\textbf{MS}} && \multicolumn{3}{c}{\textbf{D16 val}} && \multicolumn{3}{c}{\textbf{D17 val}} \\
		~ & ~ && $\mathcal{J\&F}$ & $\mathcal{J}$ & $\mathcal{F}$ && $\mathcal{J\&F}$ & $\mathcal{J}$ & $\mathcal{F}$\\
		\midrule
		CFBI~\cite{yang2020collaborative} & \Checkmark && 90.7 & 89.6 & 91.7 && 83.3 & 80.5 & 86.0\\
		XMem~\cite{cheng2022xmem}  & \Checkmark && 92.7 & 92.0 & 93.5 && 88.2 & 85.4 & 91.0 \\
		Ours  & \XSolidBrush && 92.6 & 91.5 & 93.7 && 87.1 & 83.7 & 90.5 \\
		Ours  & \Checkmark && 92.9 & 92.2 & 93.6 && 88.6 & 85.8 & 91.4 \\
		Ours$^{*}$  & \XSolidBrush && 92.8 & 91.8 & 93.8 && 88.2 & 84.5 & 91.9 \\
		Ours$^{*}$  & \Checkmark && 93.4 & 92.5 & 94.2 && 89.8 & 86.7 & 93.0 \\
		\bottomrule
	\end{tabular*}
	\caption{Results on DAVIS 2017 validation and YouTube-VOS validation split with different training data. D: DAVIS 2017, Y: YouTube 2019, S: static images, B: BL30K.  $\ddagger$ denotes pretraining on the combined DAVIS and YouTube-VOS data. }
	\label{tab:ms}
\end{table}

The results in Table~\ref{tab:ms} imply that multi-scale inference improves the performance of \system by 0.3\% and 1.7\% in terms of $\mathcal{J\&F}$ on DAVIS 2016 / 2017 validation split, and \system still outperforms existing methods.

\section{Results on Long video datasets}
In order to further evaluate the long-term performance of \system, we additionally test our method on the Long-time Video dataset~\cite{liang2020video}, which contains three videos with more than 7,000 frames in total for validation. Considering the video duration is longer and the target object(s) will undergo distinct appearance deformation or scale variations, we set the maximum memory size to 64 during inference. The comparison results are shown in Table~\ref{tab:long}.

\begin{table}[!ht]
	\centering
	\setlength{\tabcolsep}{0pt} 
	\begin{tabular*}{0.9\linewidth}{@{\extracolsep{\fill}}l | ccc }
		\toprule
		\multirow{2}{1in}{\textbf{Method}} & \multicolumn{3}{c}{\textbf{Long-time Video}}  \\
		~ & $\mathcal{J\&F}$ & $\mathcal{J}$ & $\mathcal{F}$ \\
		\midrule
		RMNet~\cite{xie2021efficient} & 59.8 & 59.7 & 60.0 \\
		JOINT~\cite{mao2021joint} & 67.1 & 64.5 & 69.6 \\
		STM~\cite{oh2019video} & 80.6 & 79.9 & 81.3 \\ 
		HMMN~\cite{shi2015hierarchical} & 81.5 & 79.9 & 83.0 \\
		STCN~\cite{cheng2021stcn} & 87.3 & 85.4 & 89.2  \\
		AOT~\cite{yang2021associating} & 84.3 & 83.2 & 85.4 \\
		AFB-URR~\cite{liang2020video} & 83.7 & 82.9 &  84.5 \\
		XMem~\cite{cheng2022xmem} & 89.8 & 88.0 & 91.6 \\
		Ours & \textbf{90.0} & \textbf{88.3} & \textbf{91.7} \\
		\bottomrule
	\end{tabular*}
	\caption{Results on the Long-time Video dataset~\cite{liang2020video}. }
	\label{tab:long}
\end{table}

We can observe that \system again achieves the best segmentation results measured in different metrics. It is worth mentioning that \system beats the methods specifically designed for long videos, \eg, AFB-URR~\cite{liang2020video} and XMem~\cite{cheng2022xmem}. We believe the performance gain is resulted from taking advantage of the instance information in query frame to facilitate the semantic matching.

\section{More Visualizations}

\begin{figure}[!ht]
	\centering
	\includegraphics[width=\linewidth]{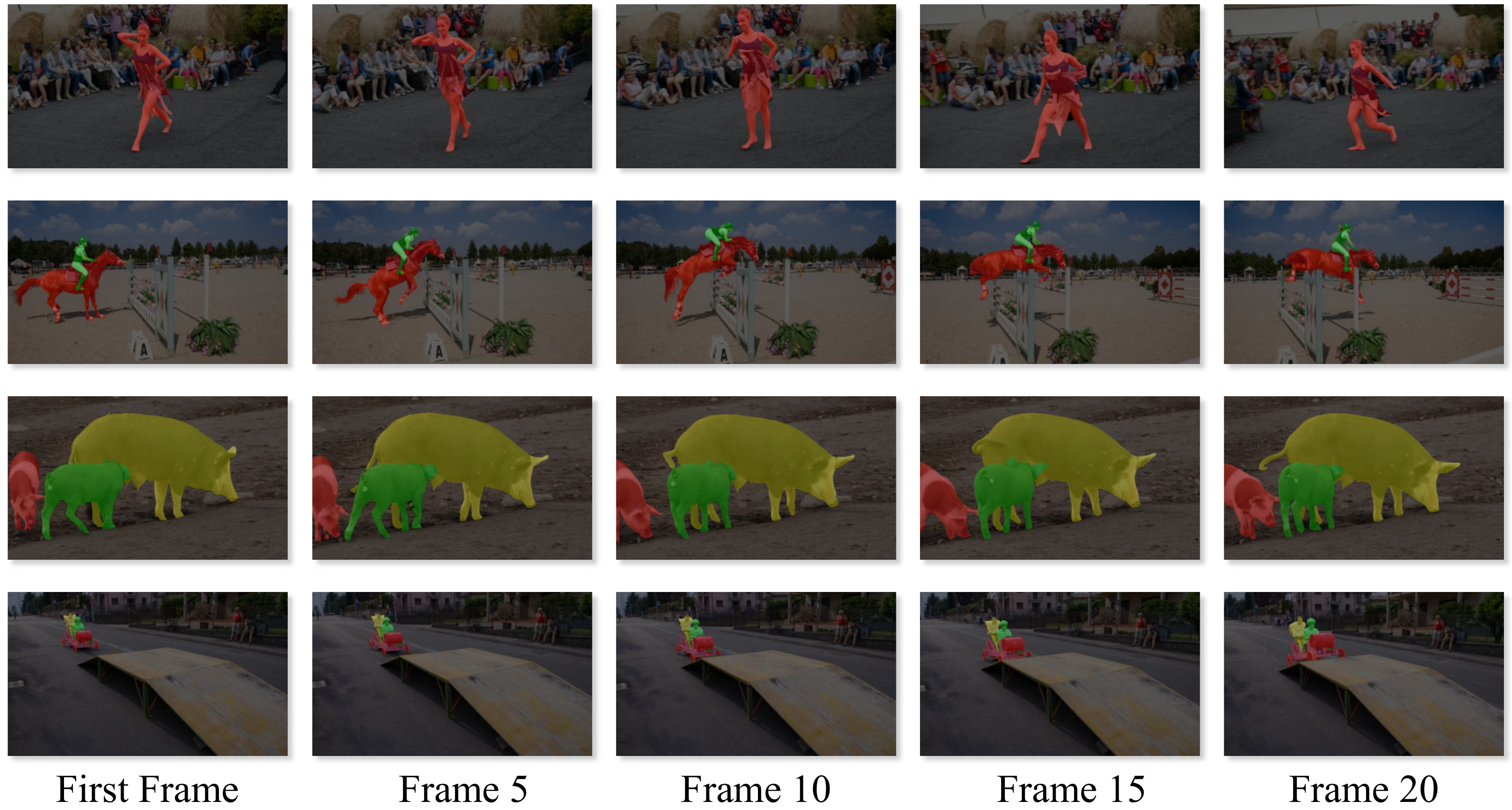}
	\vspace{-0.25in}
	\caption{Segmentation results on DAVIS 2017 validation split.}
	\label{fig:visualization1}
\end{figure}

\begin{figure}[!ht]
	\centering
	\includegraphics[width=\linewidth]{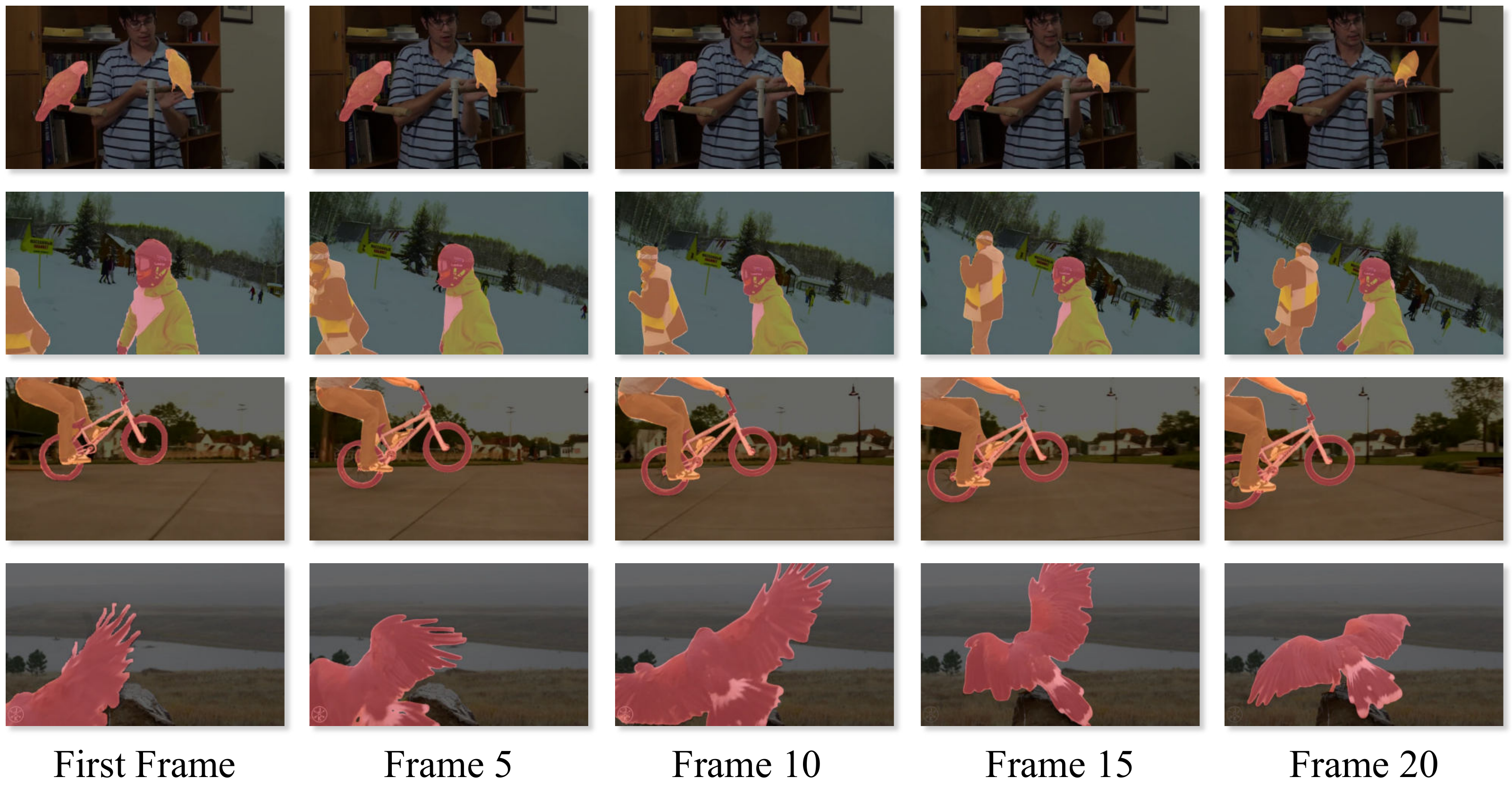}
	\vspace{-0.25in}
	\caption{Segmentation results on YouTube-VOS 2018 validation split.}
	\label{fig:visualization2}
\end{figure}

\begin{figure}[!ht]
	\centering
	\includegraphics[width=\linewidth]{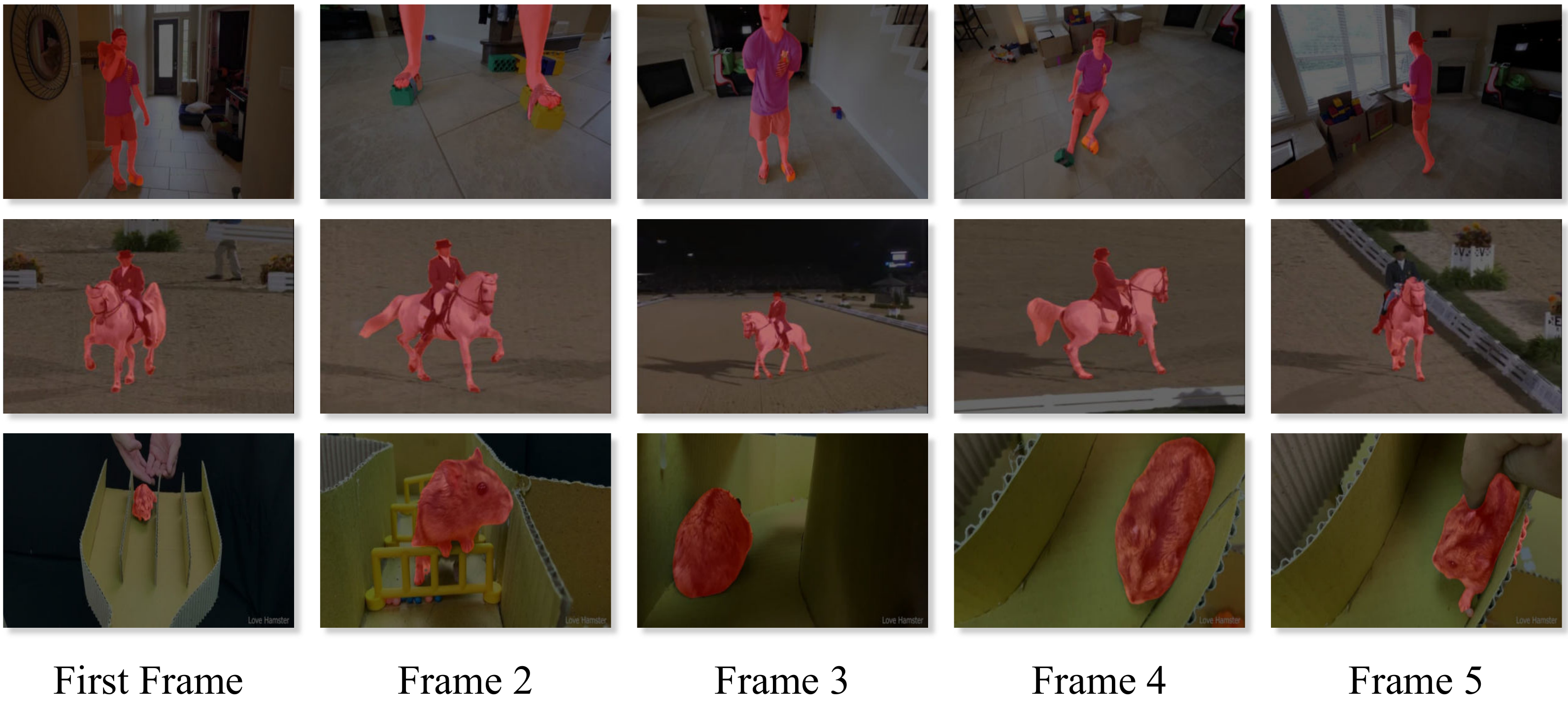}
	\vspace{-0.25in}
	\caption{Segmentation results on Long-time Video dataset.}
	\label{fig:visualization3}
\end{figure}

We show the predicted segmentation masks of \system on DAVIS 2017 val, YouTube-VOS 2018 val, and Long-time Video dataset in Figure~\ref{fig:visualization1}, Figure~\ref{fig:visualization2}, Figure~\ref{fig:visualization3}, respectively. For the short video datasets, \ie, DAVIS and YouTube-VOS, the time interval is 5, while for the long video dataset, \ie, Long-time Video dataset, the time interval is 1 since it is sparsely annotated. We can see that our method could generate accurate masks even for the objects with remarkable appearance variations.

\end{document}